Original Paper

# The Classification of Abnormal Hand Movement to Aid in Autism Detection: Machine Learning Study

Anish Lakkapragada[1]; Aaron Kline[1], BS; Onur Cezmi Mutlu[2], MS; Kelley Paskov[3], MS; Brianna Chrisman[4], MS; Nathaniel Stockham[5], MS; Peter Washington[6], PhD; Dennis Paul Wall[3], PhD

[1]Division of Systems Medicine, Department of Pediatrics, Stanford University, Stanford, CA, United States
[2]Department of Electrical Engineering, Stanford University, Stanford, CA, United States
[3]Department of Biomedical Data Science, Stanford University, Stanford, CA, United States
[4]Department of Bioengineering, Stanford University, Stanford, CA, United States
[5]Department of Neuroscience, Stanford University, Stanford, CA, United States
[6]Information and Computer Sciences, University of Hawai'i at Mānoa, Honolulu, HI, United States

**Corresponding Author:**
Peter Washington, PhD
Information and Computer Sciences
University of Hawai'i at Mānoa
2500 Campus Rd
Honolulu, HI, 96822
United States
Phone: 1 5126800926
Email: peter.y.washington@hawaii.edu

## Abstract

**Background:** A formal autism diagnosis can be an inefficient and lengthy process. Families may wait several months or longer before receiving a diagnosis for their child despite evidence that earlier intervention leads to better treatment outcomes. Digital technologies that detect the presence of behaviors related to autism can scale access to pediatric diagnoses. A strong indicator of the presence of autism is self-stimulatory behaviors such as hand flapping.

**Objective:** This study aims to demonstrate the feasibility of deep learning technologies for the detection of hand flapping from unstructured home videos as a first step toward validation of whether statistical models coupled with digital technologies can be leveraged to aid in the automatic behavioral analysis of autism. To support the widespread sharing of such home videos, we explored privacy-preserving modifications to the input space via conversion of each video to hand landmark coordinates and measured the performance of corresponding time series classifiers.

**Methods:** We used the Self-Stimulatory Behavior Dataset (SSBD) that contains 75 videos of hand flapping, head banging, and spinning exhibited by children. From this data set, we extracted 100 hand flapping videos and 100 control videos, each between 2 to 5 seconds in duration. We evaluated five separate feature representations: four privacy-preserved subsets of hand landmarks detected by MediaPipe and one feature representation obtained from the output of the penultimate layer of a MobileNetV2 model fine-tuned on the SSBD. We fed these feature vectors into a long short-term memory network that predicted the presence of hand flapping in each video clip.

**Results:** The highest-performing model used MobileNetV2 to extract features and achieved a test F1 score of 84 (SD 3.7; precision 89.6, SD 4.3 and recall 80.4, SD 6) using 5-fold cross-validation for 100 random seeds on the SSBD data (500 total distinct folds). Of the models we trained on privacy-preserved data, the model trained with all hand landmarks reached an F1 score of 66.6 (SD 3.35). Another such model trained with a select 6 landmarks reached an F1 score of 68.3 (SD 3.6). A privacy-preserved model trained using a single landmark at the base of the hands and a model trained with the average of the locations of all the hand landmarks reached an F1 score of 64.9 (SD 6.5) and 64.2 (SD 6.8), respectively.

**Conclusions:** We created five lightweight neural networks that can detect hand flapping from unstructured videos. Training a long short-term memory network with convolutional feature vectors outperformed training with feature vectors of hand coordinates and used almost 900,000 fewer model parameters. This study provides the first step toward developing precise deep learning methods for activity detection of autism-related behaviors.

## Introduction

Autism affects almost 1 in 44 people in America [1] and is the fastest growing developmental delay in the United States [2,3]. Although autism can be identified accurately by 24 months of age [4,5], the average age of diagnosis is slightly below 4.5 years [6]. This is problematic because earlier intervention leads to improved treatment outcomes [7]. Mobile digital diagnostics and therapeutics can help bridge this gap by providing scalable and accessible services to underserved populations lacking access to care. The use of digital and mobile therapies to support children with autism has been explored and validated in wearable devices [8-15] and smartphones [16-22] enhanced by machine learning models to help automate and streamline the therapeutic process.

Mobile diagnostic efforts for autism using machine learning have been explored in prior literature. Autism can be classified with high performance using 10 or fewer behavioral features [23-28]. While some untrained humans can reliably distinguish these behavioral features [25,29-36], an eventual goal is to move away from human-in-the-loop solutions toward automated and privacy-preserving diagnostic solutions [37,38]. Preliminary efforts in this space have included automated detection of autism-related behaviors such as head banging [39], emotion evocation [40-42], and eye gaze [43].

Restrictive and repetitive movement such as hand stimming is a primary behavioral feature used by diagnostic instruments for autism [44]. Because computer vision classifiers for abnormal hand movement do not currently exist, at least in the public domain, we strived to create a classifier that can detect this autism-related feature as a first step toward automated clinical support systems for developmental delays like autism.

Pose estimation and activity recognition have been explored as a method for detection of self-stimulatory behaviors. Vyas et al [45] retrained a 2D Mask region-based convolutional neural network (R-CNN) [46] to obtain the coordinates of 15 body landmarks that were then transformed into a Pose Motion (PoTion) representation [47] and fed to a convolutional neural network (CNN) model for a prediction of autism-related atypical movements. This approach resulted in a 72.4% classification accuracy with 72% precision and 92% recall. Rajagopalan and Goecke [48] used the Histogram of Dominant Motions (HDM) representation to train a model to detect self-stimulatory behaviors [48]. On the Self-Stimulatory Behavior Dataset (SSBD) [49], which we also used in this study, the authors achieved 86.6% binary accuracy when distinguishing head banging versus spinning and 76.3% accuracy on the 3-way task of distinguishing head banging, spinning, and hand flapping. We note that they did not train a classifier with a control class absent of any self-stimulatory behavior. Zhao et al [50] used head rotation range and rotations per minute in the yaw, pitch, and roll directions as features for autism detection classifiers. This reached 92.11% classification accuracy with a decision tree model that used the head rotation range in the roll direction and the amount of rotations per minute in the yaw direction as features.

Building upon these prior efforts, we developed a computer vision classifier for abnormal hand movement displayed by children. In contrast to prior approaches to movement-based detection of autism, which use extracted activity features to train a classifier to detect autism directly, we aim to detect autism-related behaviors that may contribute to an autism diagnosis but that may also be related to other behavioral symptoms. We trained our abnormal hand movement classifier on the SSBD, as it is the only publicly available data set of videos depicting abnormal hand movement in children. We used cross-validation and achieved an F1 score of 84% using convolutional features emitted per frame by a fine-tuned MobileNetV2 model fed into a long short-term memory (LSTM). We also explored privacy-preserving hand-engineered feature representations that may support the widespread sharing of home videos.

## Methods

### Overview

We compared five separate training approaches: four subsets of MediaPipe hand landmarks fed into an LSTM and fine-tuned MobileNetV2 convolutional features fed into an LSTM. The hand landmark approaches provided an exploration of activity detection on privacy-preserved feature representations. Because we strived to use machine learning classifiers in low-resource settings such as mobile devices, we additionally aimed to make our models and feature representations as light as possible.

### Data Set

We used the SSBD [49] for training and testing of our models. To the best of our knowledge, SSBD is the only publicly available data set of self-stimulatory behaviors containing examples of head banging, hand flapping, and spinning. SSBD includes the URLs of 75 YouTube videos, and for each video, annotations of the time periods (eg, second 1 to second 35) when each self-stimulatory behavior was performed. Multiple videos contain multiple time periods for the same behavior (eg, seconds 1-3 and 5-9 both contain hand flapping) as well as multiple behaviors (eg, seconds 1-3 show head banging and seconds 5-9 show hand flapping). We only used the hand flapping annotations.

### Preprocessing

To obtain control videos absent of hand flapping displays, we first downloaded all YouTube videos in SSBD that contained sections of hand flapping. Each section in a video exhibiting hand flapping was extracted to create a new clip. The parts of the video without hand flapping (ie, with no annotations) were isolated to create control clips. This data curation process is illustrated in Figure 1.

After extracting all positive and control clips from the downloaded videos, we aimed to maximize the amount of training data in each class. Because a hand flapping event occurs within a couple of seconds, we split any clips longer than 2 seconds into smaller clips. We manually deleted any videos that were qualitatively shaky or of low quality. In total, we extracted 50 video clips displaying hand flapping and 50 control videos.

**Figure 1.** Extraction of positive and control videos. Sections of a video demonstrating hand flapping are separated to create positive videos, and segments between the hand flapping sections are used as control videos.

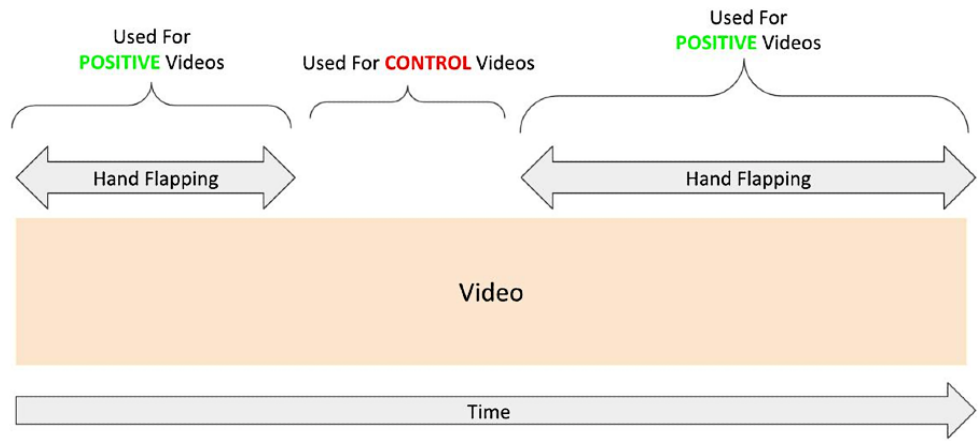

## Feature Extraction

We evaluated five separate feature extraction methods. For four of them, we used the numerical coordinates of the detected hand landmarks concatenated into a 1-dimensional vector as the primary feature representation. For the remaining model, we fine-tuned a mobile-optimized CNN, MobileNetV2 [51], to learn features derived from raw image sequences. We noted that the landmark-based feature representations are privacy-preserved, as they do not require the face of the participant to be shown in the given data for adequate classification.

To extract the hand coordinates, we used MediaPipe, a framework hosted by Google that detects the landmarks on a person's face, hands, and body [52]. MediaPipe's hand landmark detection model provides the (x, y, z) coordinates of each of the 21 landmarks it detects on each hand. The x coordinate and y coordinate describe how far the landmark is on the horizontal and vertical dimensions, respectively. The z coordinate provides an estimation of how far the landmark is from the camera. When MediaPipe does not detect a landmark, the (x, y, z) coordinates are all set to 0 for that landmark.

The first landmark-based feature representation approach we tried used all 21 landmarks on each hand provided by MediaPipe to create the location vector fed into the LSTM. SSBD's videos mostly contain children whose detected hand landmarks are closer together due to smaller hands. This could be a problem when generalizing to older individuals with wider gaps between hand landmarks. To help the model generalize beyond hand shape, one possible solution is to use a curated subset of landmarks.

To eliminate hand shape all together, one could use only one landmark. We tried this method by using a single landmark at the base of the hand. However, because the videos in SSBD may be shaky, reliance on MediaPipe being able to detect this landmark may have led to empty features for some frames. One way to circumvent this problem is to take the mean of all the (x, y, z) coordinates of detected landmarks and use the average coordinate for each hand. We call this method the "mean landmark" approach.

We took the first 90 frames of a video and for each frame, we concatenated the feature vectors and used them as input for each timestep of an LSTM model (Figure 2). We experimented with subsets of landmarks provided by MediaPipe; we tried using all 21 landmarks, 6 landmarks (5 at each fingertip and 1 at the base of the hand), and with single landmarks. We note that the concatenated coordinates of landmarks will always form a vector that is 6 times larger than the number of landmarks used because there are 3 coordinates for a single landmark and 2 hands for which each landmark can be detected.

**Figure 2.** Hand flapping detection workflow. The initial 90 frames of a single video are each converted to a feature vector, consisting of either the location of coordinates as detected by MediaPipe (depicted here) or a feature vector extracted from the convolutional layers of a MobileNetV2 model. For all feature extraction methods, the resulting feature vectors are passed into an LSTM. The LSTM's output on the final timestep is fed into a multilayer perceptron layer to provide a final binary prediction. LSTM: long short-term memory.

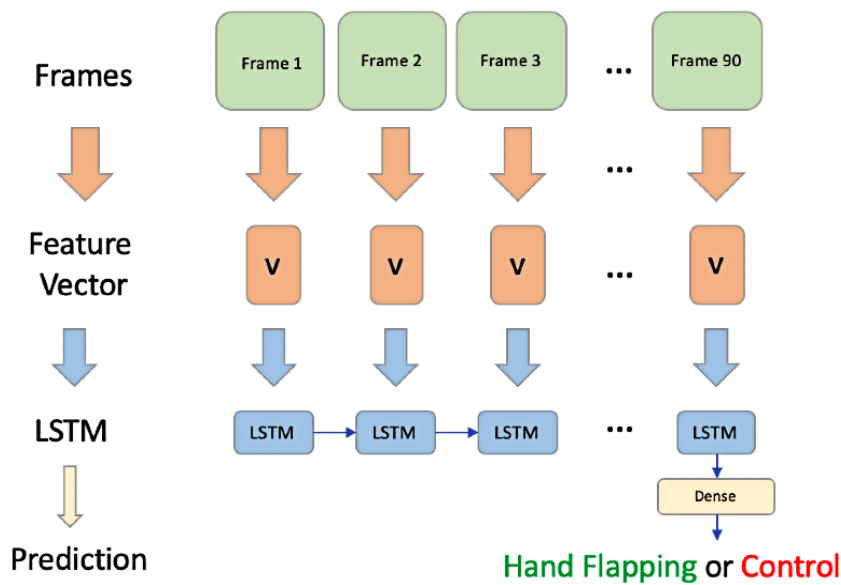

### Model Architecture

The neural network architecture we used for all experiments consisted of an LSTM layer with a 64-dimensional output. The output of the LSTM was passed into a fully connected layer with sigmoid activation to obtain a binary prediction. To minimize overfitting, we also inserted a dropout layer between the LSTM and the dense layer with a dropout rate of 30%. The landmark-based models contained nearly 3 million parameters. (Table 1). We note that the number of parameters depends on the feature approach; Table 1 shows the number of parameters based on our heaviest feature approach of using all 21 landmarks.

We experimented with other model architectures before selecting this model. We found that adding more than one LSTM or fully connected layer did not cause any notable difference in performance; thus, we removed these layers to minimize the model's capacity for overfitting. We also experimented with the output dimensionality of the LSTM; we tried 8, 16, 32, and 64. We found that using 32 and 64 performed similarly, with 64 usually performing slightly better.

**Table 1.** Number of parameters in the neural networks using hand landmarks as features. The two feature extraction models collectively contained 3,133,336 parameters. By contrast, MobileNetV2 feature extraction contained 2,260,546 parameters with 2 output classes.

| Layer | Parameters, n |
| --- | --- |
| MediaPipe Hand Detector | 1,757,766 |
| MediaPipe Landmark Extractor | 1,375,570 |
| LSTM[a] (64 units) | 48,896 |
| Dropout (30%) | 0 |
| Dense | 65 |
| Total | 3,182,297 |

[a]LSTM: long short-term memory.

### Model Training

We trained all models with binary cross-entropy loss using Adam optimization [53]. We tried learning rates of 0.0005, 0.0001, 0.0005, 0.001, and 0.1, and found that in almost all cases 0.01 worked best. All models and augmentations were written using Keras [54] with a TensorFlow [55] back end run on Jupyter. No GPUs or specialized hardware were required due to the low-dimensional feature representation, and training a single model took a few minutes on a CPU with 32GB of RAM.

For all models, we trained the model until there was consistent convergence for 10 or more epochs. This resulted in 75 epochs of training across all models. After training, we reverted the model's weights to its weights for which it performed best. We used this strategy for all feature approaches.

## Results

### Overview

We used 5-fold cross validation to evaluate each model's average accuracy, precision, recall, and F1 score across all folds for training and testing. However, because of our small data set, the particular arrangement of the videos in each fold substantially affected the model's performance. To minimize this effect, we ran the 5-fold cross-validation procedure 100 times, each with a different random seed, resulting in a total of 500 distinct folds. We further ensured that each fold was completely balanced in both the training and testing set (50% head banging and 50% not head banging). In all folds, there were 10 videos displaying hand flapping and 10 videos displaying head banging.

We report the mean and SD of each metric across all 500 folds as well as the area under receiver operating characteristics (AUROC). For all feature approaches, we also show the average receiver operating characteristics (ROC) curve across all folds.

### All Hand Landmarks

This approach used all 21 landmarks on both hands for a total of 42 unique landmarks. We show the results of this approach in Table 2. In Figure 3, we show the ROC curves of the model with and without augmentations.

When using all the landmarks, we used graphical interpolation to fill in the coordinates of missing landmarks to help reduce the effects of camera instability. However, when we tried this, we found that it often decreased accuracy and resulted in higher SDs. We therefore decided to discontinue using interpolation when evaluating the approaches described in the next section. We conjecture that the inability of MediaPipe to detect hand key points could be a salient feature for hand flapping detection, and this feature becomes obfuscated once key points are interpolated.

Table 2. Model performance for training and testing when using all hand landmarks in the feature representation.

| Run type | Accuracy (SD; %) | Precision (SD; %) | Recall (SD; %) | F1 (SD; %) |
|---|---|---|---|---|
| Training | 79.7 (1.6) | 82.4 (2.67) | 76.5 (3.0) | 79.0 (1.7) |
| Testing | 68.0 (2.66) | 70.3 (3.6) | 65.34 (5.0) | 66.6 (3.35) |

Figure 3. Receiver Operating Characteristics (ROC) curve across all runs when using all hand landmarks. We achieved an area under receiver operating characteristics of 0.748 (SD 0.26).

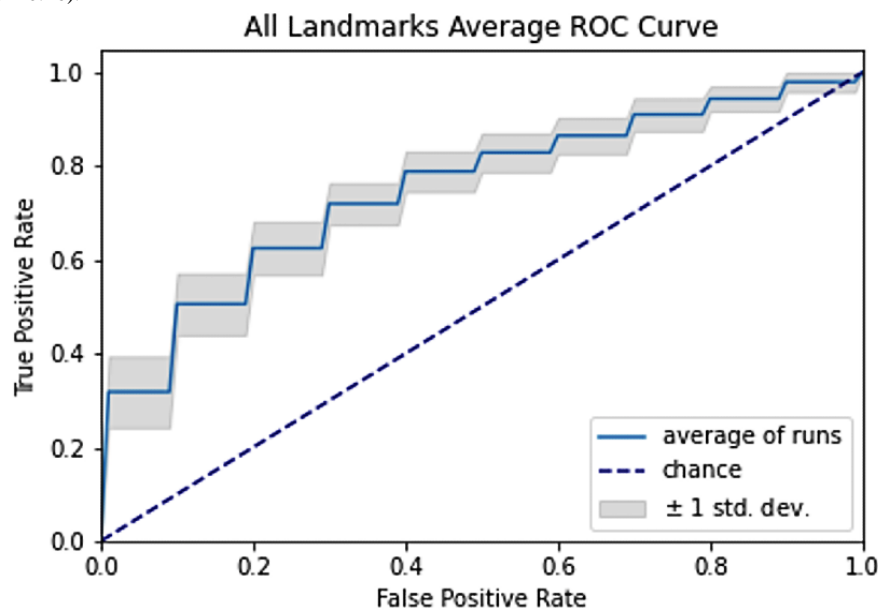

### Single Hand Landmark

Here, we describe the mean and one landmark approaches, both of which relied on a single landmark on each hand as the feature representation. We show the results of both approaches, with and without augmentations, in Table 3. In Figure 4, we show the average ROC curve for both approaches.

Table 3. Model performance for mean versus single landmark feature representations with and without data augmentation.

| Approach | Train/test | Accuracy (SD; %) | Precision (SD; %) | Recall (SD; %) | F1 (SD; %) |
|---|---|---|---|---|---|
| Mean landmark | Training | 69.2 (4.1) | 70.4 (5.3) | 70.6 (7.0) | 68.9 (5.12) |
| Mean landmark | Testing | 65.5 (4.5) | 66.7 (7.4) | 66.9 (9.6) | 64.2 (6.8) |
| One landmark | Training | 69.2 (3.4) | 70.47 (4.4) | 69.71 (6.7) | 68.7 (4.4) |
| One landmark | Testing | 65.8 (4.3) | 66.5 (7.5) | 68.0 (6.7) | 64.9 (6.5) |

XSL•FO
RenderX

**Figure 4.** Average ROC curve for the mean (left plot) and one (right plot) landmark approach. The mean landmark approach yielded an area under receiver operating characteristics (AUROC) of 0.73 (SD 0.04), and the one landmark approach yielded an AUROC of 0.751 (SD 0.03). ROC: receiver operating characteristics.

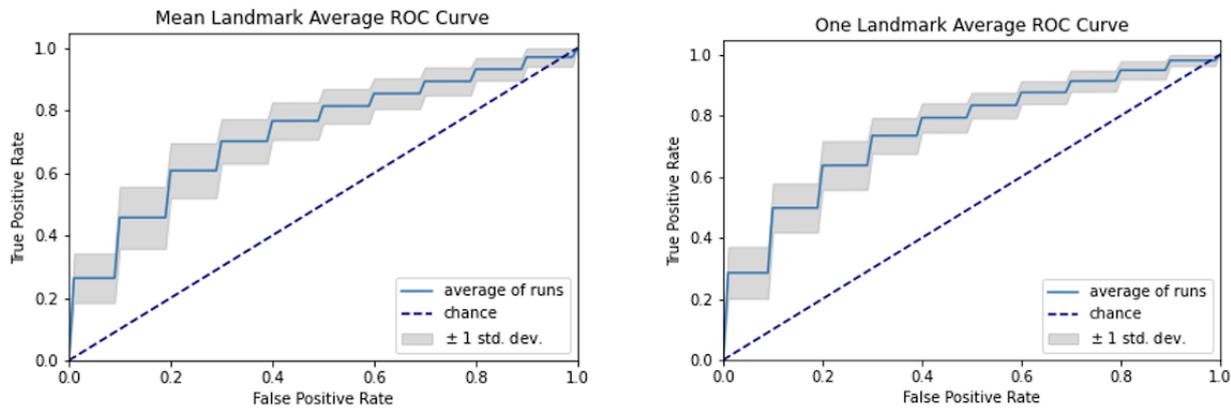

### Six Hand Landmarks

We used the six landmarks on the edges of the hands to create the location frames. We achieved an F1 score and classification accuracy of about 72.3% (Table 4). We also achieved an AUROC of 0.76 (Figure 5).

Of all of the landmark-based approaches, the six landmarks approach yielded optimal results. All of the validation metrics were higher with this approach than those previously discussed.

**Table 4.** Model performance in training and testing for feature representations containing six landmarks.

| Run type | Accuracy (SD; %) | Precision (SD; %) | Recall (SD; %) | F1 (SD; %) |
| --- | --- | --- | --- | --- |
| Training | 76.8 (1.95) | 78.7 (2.9) | 74.7 (3.5) | 76.2 (2.1) |
| Testing | 69.55 (2.7) | 71.7 (3.5) | 67.5 (5.5) | 68.3 (3.6) |

**Figure 5.** Receiver Operating Characteristics (ROC) curve for the six landmarks approach across all runs. We achieved an area under receiver operating characteristics of 0.76 (SD 0.027) with this approach.

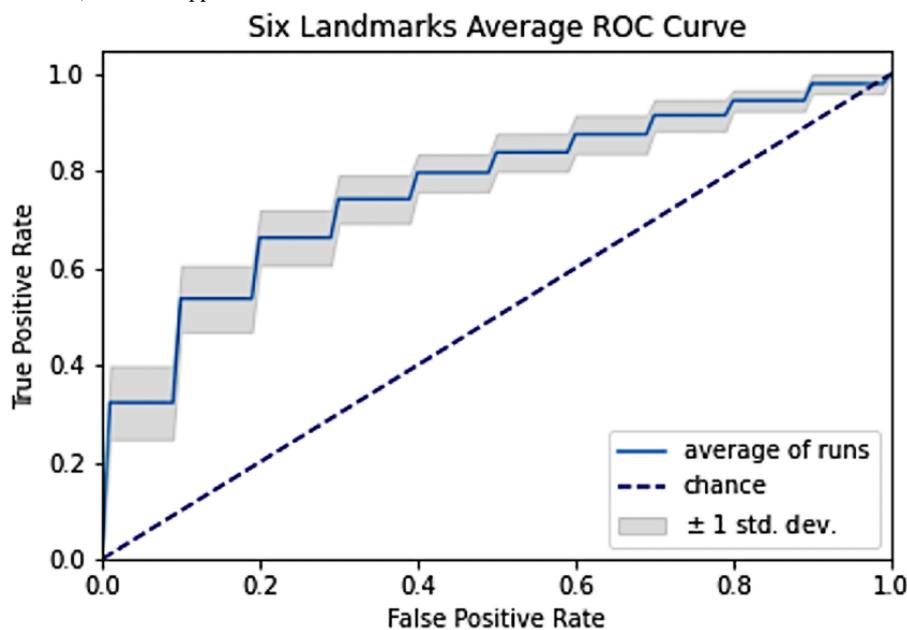

### MobileNetV2 Model

In the approaches discussed so far, MediaPipe was consistently used as a feature extractor to bring each video frame into a lower-dimensional vector representation. Here, we substituted the MediaPipe feature extractor with MobileNetV2's [51] convolutional layers (pretrained on ImageNet [56] and fine-tuned on SSBD) as a feature extractor. As with the landmark-based approaches, this extracted vector was fed into an LSTM network to obtain a prediction for whether hand flapping was present in the video. We evaluated this model on the same 100 data sets (500 total folds), as we used for all other approaches. The ROC curve of this model is shown in Figure 6, and the metrics are detailed in Table 5.

The MobileNetV2 model achieved an accuracy and F1 score both around 85%, surpassing the performance of all the landmark-based approaches. The MobileNetV2 models also had a higher capacity to overfit, achieving near perfect accuracies in training (>99.999%), whereas all landmark-based approaches never surpassed 90% for any of the training metrics. We conjecture that this is because the MobileNet V2 model has learned both the feature extraction and discriminative steps of the supervised learning process.

**Figure 6.** Receiver Operating Characteristics (ROC) curve of the Mobile Net. With this method, we achieved an area under receiver operating characteristics of 0.85 (SD 0.03).

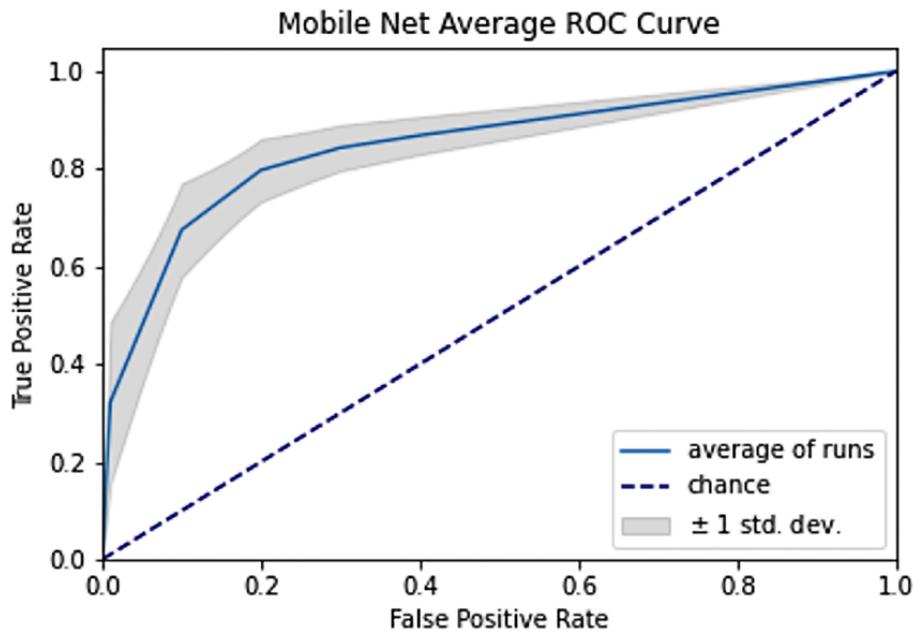

**Table 5.** Model performance in training and testing when using MobileNetV2 convolutional layers as the feature extractor.

| Run type | Accuracy (SD; %) | Precision (SD; %) | Recall (SD; %) | F1 (SD; %) |
| --- | --- | --- | --- | --- |
| Training | 97.7 (1.0) | 99.5 (0.0) | 95.9 (1.7) | 97.6 (1.0) |
| Testing | 85.0 (3.14) | 89.6 (4.3) | 80.4 (6.0) | 84.0 (3.7) |

## Comparison of Results

We conducted a 2-sided $t$ test to determine whether the differences we observed for each approach (including the MobileNetV2 method) were statistically significant. We applied Bonferroni correction across the comparisons, deeming a $P$ value <.005 as statistically significant. We show the $P$ values from comparing all the approaches with each other on the 4 aforementioned metrics in Table 6.

Most of the comparisons between approaches were statistically significant after Bonferroni correction. The two single landmark approaches (mean and one landmark) were not statistically significant for any of the metrics.

**Table 6.** We conducted a 2-sided t test to determine whether the differences in results for each approach were statistically significant. We display $P$ values for the 500 accuracy, precision, recall, and F1 values.

|  | All landmarks vs mean landmark ($P$ value) | All landmarks vs one landmark ($P$ value) | All landmarks vs six landmarks ($P$ value) | All landmarks vs mobile net ($P$ value) | Six landmarks vs mean landmark ($P$ value) | Six landmarks vs one landmark ($P$ value) | Six landmarks vs mobile net ($P$ value) | Mean landmark vs one landmark ($P$ value) | Mean landmark vs mobile net ($P$ value) | One landmark vs mobile net ($P$ value) |
| --- | --- | --- | --- | --- | --- | --- | --- | --- | --- | --- |
| Accuracy | <.001 | <.001 | <.001 | <.001 | <.001 | <.001 | <.001 | .67 | <.001 | <.001 |
| Precision | <.001 | <.001 | .007 | <.001 | <.001 | <.001 | <.001 | .85 | <.001 | <.001 |
| Recall | .15 | .01 | .004 | <.001 | .59 | .66 | <.001 | .42 | <.001 | <.001 |
| F1 | .002 | .02 | .001 | <.001 | <.001 | <.001 | <.001 | .50 | <.001 | <.001 |

## Discussion

### Principal Results

We explored several feature representations for lightweight hand flapping classifiers that achieved respectable performance on the SSBD. The highest-performing model used MobileNetV2 to extract features and achieved a test F1 score of 84 (SD 3.7). A model trained with all hand landmarks reached an F1 score of 66.6 (SD 3.35). A model trained with a select 6 landmarks reached an F1 score of 68.3 (SD 3.6). A model trained using a single landmark at the base of the hands reached an F1 score of 64.9 (SD 6.5).

One point of interest in this study is the trade-off between privacy-preserved solutions and performance in diagnostic machine learning tasks. While the MobileNetV2 model outperformed all the MediaPipe classifiers, the MobileNetV2 model lacks the capability to preserve the privacy of the participants, as the participants' faces were ultimately used in the data needed for classification. We expect this to be a difficulty for future research in the behavioral diagnostic space.

### Limitations

The primary limitation of this approach is that without further class labels across a variety of hand-related activities and data sets, there is a probable lack of specificity in this model when generalizing to other data sets beyond the SSBD. Hands can move but not display hand flapping or self-stimulatory movement. Furthermore, stereotypic use of hands may occur in the absence of a formal autism diagnosis. Multi-class models that can distinguish hand movement patterns are required for this degree of precision. Such models cannot be built without corresponding labeled data sets, and we therefore highlight the need for the curation of data sets displaying behaviors related to developmental health care.

For this study to truly generalize, further validation is required on data sets beyond the SSBD. While the SSBD was curated with autism diagnosis in mind, the paper describing the original data set does not necessarily include children with confirmed autism diagnoses. Existing mobile therapies that collect structured videos of children with autism [16-18,40] can be used to acquire data sets to train more advanced models, and these updated models can be integrated back into the digital therapy to provide real-time feedback and adaptive experiences.

### Opportunities for Future Work

There are myriad challenges and opportunities for computer vision recognition of complex social human behaviors [57], including socially motivated hand mannerisms. Additional prospects for future work include alternative feature representation and incorporation of modern architectures such as transformers and other attention-based models.

The hand movement classifier we describe here is one of a potential cocktail of classifiers that could be used in conjunction not only to extract features relevant to an autism diagnosis but also to provide insight into which particular symptoms of autism a child is exhibiting. The primary benefit of this approach is for greater explainability in medical diagnoses and a strive toward specificity in automated diagnostic efforts.

### Comparison With Prior Work

#### Gaze Patterns

Gaze patterns often differ between autism cases and controls. Chang et al [58] found that people with autism spend more time looking at a distracting toy than a person engaging in social behavior in a movie when compared to those with typical development. This demonstrated that gaze patterns and a preference to social stimuli is an indicator of autism. Gaze patterns have been used as a feature in machine learning classifiers. Jiang et al [59] created a random forest classifier that used as an input a participant's performance in classifying emotions and other features about their gaze and face. They achieved an 86% accuracy for classifying autism with this approach. Liaquat et al [60] used CNNs [61] and LSTMs on a data set of gaze patterns and achieved a 60% accuracy on classifying autism.

#### Facial Expression

Another behavior feature relevant to autism detection is facial expression. Children with autism often evoke emotions differently than neurotypical peers. Volker et al [62] found that typically developing raters had more difficulty with recognizing sadness in the facial expressions of those with autism than controls. This finding was confirmed by Manfredonia et al [20] who used an automated facial recognition software to compare how easily those with autism and those who are neurotypical could express an emotion when asked. They found that people with autism had a harder time producing the correct facial expression when prompted compared to controls. People with autism typically have less facial symmetry [63]. Li et al [64] achieved an F1 score of 76% by using a CNN to extract features of facial expressions in images that were then used to classify autism. CNNs, along with recurrent neural networks [65], were also applied in Zunino et al's [66] work where videos were used to classify autism. They achieved 72% accuracy on classifying those with autism and 77% accuracy on classifying typically developing controls.

#### On-Body Devices

Smartwatch-based systems and sensors have been used to detect repetitive behaviors to aid intervention for people with autism. Westeyn et al [67] used a hidden Markov model to detect 7 different stimming patterns using accelerometer data. They reached a 69% accuracy with this approach. Albinali et al [68] tried using accelerometers on the wrists and torsos to detect stimming in people with autism. They achieved an accuracy of 88.6%. Sarker et al [69] used a commercially available smartwatch to collect data of adults performing stimming behaviors like head banging, hand flapping, and repetitive dropping. They used 70 features from accelerometer and gyroscope data streams to build a gradient boosting model with an accuracy of 92.6% and an F1 score of 88.1%.

#### Pose Estimation

Pose estimation and activity recognition have also been used to detect self-stimulatory behaviors. Vyas et al [45] retrained a 2D Mask R-CNN [46] to get the coordinates of 15 key points that were then transformed into a PoTion representation [47] and fed into a CNN model for a prediction of autism-related behavior. This approach resulted in a 72.4% classification accuracy with 72% precision and 92% recall. We note that they used a derived 8349 episodes from private videos of the Behavior Imaging company to train their model. Rajagopalan and Goecke [48] used the HDM from a video that gives the dominant motions detected to train a discriminatory model to detect self-stimulatory behaviors. On the SSBD [49], which we also used in this study, they reached an 86.6% accuracy on distinguishing head banging versus spinning behavior and a 76.3% accuracy on distinguishing head banging, spinning, and hand flapping behavior. We note that they did not train a classifier with a control class. Another effort sought to determine whether individuals with autism nod or shake their head differently than neurotypical peers. They used head rotation range and amount of rotations per minute in the yaw, pitch, and roll directions as features for the machine learning classifiers to detect autism [50]. They achieved a 92.11% accuracy from a decision tree model that used the head rotation range in the roll direction and the amount of rotations per minute in the yaw direction as features.

## Acknowledgments

The study was supported in part by funds to DPW from the National Institutes of Health (1R01EB025025-01, 1R01LM013364-01, 1R21HD091500-01, 1R01LM013083); the National Science Foundation (Award 2014232); The Hartwell Foundation; Bill and Melinda Gates Foundation; Coulter Foundation; Lucile Packard Foundation; Auxiliaries Endowment; The Islamic Development Bank (ISDB) Transform Fund; the Weston Havens Foundation; and program grants from Stanford's Human Centered Artificial Intelligence Program, Precision Health and Integrated Diagnostics Center, Beckman Center, Bio-X Center, Predictives and Diagnostics Accelerator, Spectrum, Spark Program in Translational Research, MediaX, and the Wu Tsai Neurosciences Institute's Neuroscience:Translate Program. We also acknowledge generous support from David Orr, Imma Calvo, Bobby Dekesyer, and Peter Sullivan. PW would like to acknowledge support from Mr Schroeder and the Stanford Interdisciplinary Graduate Fellowship as the Schroeder Family Goldman Sachs Graduate Fellow.

## Conflicts of Interest

DPW is the founder of Cognoa.com. This company is developing digital health solutions for pediatric health care. AK works as part-time consultant to Cognoa.com. All other authors declare no competing interests.

## Abbreviations

**AUROC:** area under receiver operating characteristics
**CNN:** convolutional neural network
**HDM:** Histogram of Dominant Motions
**LSTM:** long short-term memory
**PoTion:** Pose Motion
**R-CNN:** region-based convolutional neural network
**ROC:** receiver operating characteristics
**SSBD:** Self-Stimulatory Behavior Dataset